\documentclass[12pt]{article}

\usepackage{epsfig,psfrag,amsmath,amssymb,latexsym}
\usepackage{amscd }
\usepackage{amsfonts}
\usepackage{graphicx}
\usepackage{mathrsfs}
\usepackage{xcolor}
\usepackage[utf8]{inputenc} % allow utf-8 input
\usepackage[T1]{fontenc}    % use 8-bit T1 fonts
\usepackage{hyperref}       % hyperlinks
\usepackage{url}            % simple URL typesetting
\usepackage{booktabs}       % professional-quality tables
\usepackage{amsfonts}       % blackboard math symbols
\usepackage{nicefrac}       % compact symbols for 1/2, etc.
\usepackage{microtype}      % microtypography
\usepackage{amsmath}        % math package
\usepackage{tikz} \usetikzlibrary{arrows,shapes}
\usepackage{wrapfig}			%embedd figures in text
\usepackage[font=small]{caption, subcaption} %add subcaption to subfigures

\newcommand{\B}{\{0,1\}}

\newcommand{\R}{\mathbb{R}}

\begin{document}

\title{Low-Rank Boolean Matrix Approximation by Integer Programming}

\author{R\'{e}ka \'{A}. Kov\'{a}cs\thanks{Mathematical Institute, Radcliffe Observatory Quarter, Woodstock Road, Oxford OX2 6GG, U.K.; \texttt{reka.kovacs@maths.ox.ac.uk}; supported by the 
Robin \& Nadine Wells Scholarship from St Cross College Oxford}, 
Oktay Gunluk\thanks{IBM Research, Yorktown Heights, NY 10598; \texttt{gunluk@us.ibm.com}},
Raphael A.\ Hauser\thanks{Mathematical Institute, Radcliffe Observatory Quarter, Woodstock Road, Oxford OX2 6GG, U.K.; Alan Turing Institute, British Library, 96 Euston Road, London NW1 2DB, U.K.; Pembroke College, St Aldates, Oxford OX1 1DW, U.K.; \texttt{hauser@maths.ox.ac.uk}; This work was supported by The Alan Turing Institute under the EPSRC grant EP/N510129/1.}
}

\maketitle

\begin{abstract}
Low-rank approximations of data matrices are an important dimensionality reduction tool in machine learning and regression analysis. We consider the case of categorical variables, where it can be formulated as the problem of finding low-rank approximations to Boolean matrices. In this paper we give what is to the best of our knowledge the first integer programming formulation that relies on only polynomially many variables and constraints, we discuss how to solve it computationally and report numerical tests on synthetic and real-world data. 
\end{abstract}

\section{Introduction}

A common problem in machine learning and regression analysis is to predict the value of an as of yet unobserved output  variable of interest as a function of $m$ observed input variables called features. The functional dependence between the input and output variables has to be learned from a set of $n$ samples, or items, each of whose inputs and outputs are known. When $n$ is small in relation to $m$, this task may be affected by overfitting \cite{hastie2013elements}, but in applications it is typically observed that the data is inherently well approximated by a lower dimensional representation, and that reducing the dimensionality dramatically reduces the overfitting. A classical technique to achieve dimensionality reduction is linear factor analysis \cite{PCA, SVDD, factor}: Given a data matrix $X\in\R^{n\times m}$ whose rows correspond to $n$ items and columns to $m$ features,  compute $C\in\R^{n\times k}$ and $R\in\R^{k\times m}$ such that the Frobenius norm $\|X-CR\|_F$ of the approximation error is minimal for some fixed rank $k\in\mathbb{N}$. The rank-$k$ approximation $CR$ describes the data matrix using only $k$ implicit features: the rows of $R$ specify how the observed variables relate to the implicit features, while the rows of $C$ show how the observed variables of each item can be (approximately) expressed as a linear combination of the $k$ implicit features.

Many practical data sets contain a mixture of different data types. In this paper we concentrate on categorical variables. For example, in the data set of congressional votes discussed in the numerical section of this paper, items correspond to $435$ members of congress, and features to votes on $16$ different bills. The voting behavior of each member can be represented by two Boolean variables per bill, a first variable taking the value $1$ if the member voted ``yes'', and a second variable that takes the value $1$ if they voted ``no''. Note that in case of abstentions, both variables take the value $0$, indicating an absence of both categorical features. 
%Under this conversion, the $16$ categorical features can be represented by $32$ Boolean features, and the entire data consists of a Boolean matrix $X$ of size $435\times 32$. 
Such an expansion of categorical variables into Boolean variables proportional to the number of different categories is both typical and necessary because of an asymmetry of treating $1$s and $0$s under Boolean arithmetic. 

For Boolean data matrices $X\in\B^{n \times m}$ it is natural to require that the factor matrices $C$ and $R$ are Boolean as well. This requirement introduces an intrinsic difficulty into the problem because real arithmetic is replaced by arithmetic over the Boolean semiring in which $1+1=1$ holds. Boolean matrix multiplication is defined as 
$X=C\circ R \iff x_{i,j}=\bigvee_{\ell=1}^k c_{i,\ell}\land r_{\ell,j}$ for some Boolean matrices $X\in\B^{n\times m}, C\in\B^{n\times k},R\in\B^{k\times m}$. Note that for Boolean vectors $a,b\in\{0,1\}^k$, it holds that $\bigvee_{\ell=1}^k a_{\ell}\land b_{\ell}=\min\{1,\sum_{\ell=1}^k a_{\ell}b_{\ell}\}$. An optimal rank-$k$ Boolean matrix approximation for $X\in\B^{n\times m}$ and $k\in\mathbb{N}$ is then given by $C\in\B^{n\times k}$ and $R\in\B^{k\times m}$ for which 
$\|X-C\circ R\|_F^2$ is minimal. The Boolean rank of $X$ is defined as the smallest $k$ for which the approximation error is zero \cite{kim1982boolean}. Note that the Boolean rank of a Boolean matrix $X$ may differ from its linear algebraic rank. In the following example, inspired by  \cite{miettinenTalk}, let 
$X=\left[\begin{smallmatrix}1&1&0\\1&1&1\\0&1&1\end{smallmatrix}\right]$ be a data matrix of Boolean variables 
$x_{i,j}$ that indicate if worker $i$ has access to room $j$. The Boolean factorization 
$X=C \circ R=\left[
\begin{smallmatrix}
1 & 0  \\
1 & 1 \\
0 & 1 \\
\end{smallmatrix}\right]
\circ
\left[\begin{smallmatrix}
1 & 1 & 0  \\
0 & 1 & 1 \end{smallmatrix}\right]$ 
is of exact Boolean rank $2$ and reveals that there are two different roles, one requiring access to rooms $1$ and $2$, and the other requiring access to rooms $2$ and $3$, and that worker $2$ serves in both roles, whereas workers $1$ and $3$ serve only in one. In contrast, treating $X$ as a real matrix renders it of linear algebraic rank $3$, and the best rank-$2$ approximation 
$X\approx\left[
\begin{smallmatrix}
1.207 & 0.707  \\
1.207 &  0  \\
1.207 & -0.707
\end{smallmatrix}\right]
\left[\begin{smallmatrix}
0.5 & 0.707 & 0.5  \\
0.707 & 0 &  - 0.707\end{smallmatrix}\right]$
fails to reveal a clear interpretation. 

Interpreting $X$ as the node-node incidence matrix of a bipartite graph $G$, the problem of finding the Boolean rank of $X$ has an interpretation as a minimum edge covering of $G$ by bi-cliques, which is a well known NP-complete problem \cite{orlin}. Correspondingly finding the best Boolean rank-$k$ approximation of $X$ has an interpretation of minimizing the number of errors in approximate coverings of $G$ by $k$ bi-cliques. 

Boolean rank-$k$ approximation is a problem that is generally solved via heuristics that may \cite{frolov1, frolov2} or may not yield a Boolean factorization \cite{leeuw, LPCA, binaryPCA, udell}. The method developed in \cite{formcon2, formcon1} can be used to find exact Boolean rank-$k$ decompositions, but not approximations. The method of \cite{tiling} produces Boolean rank-$k$ approximations but treat the errors in $1$ and $0$ asymmetrically. \cite{Miettinen} presents a powerful heuristic for the correct error term and returns genuinely Boolean factors. Building on methods of \cite{Vaidya:2006}, the authors of \cite{Lu:2008} presented an integer programming model that is closely related to low-rank Boolean matrix approximation but relies on mining an initial set of patterns from which the approximating factors are composed. They also provide an integer programming model with an exponential number of variables and constraints for low-rank Boolean matrix approximation, but do not detail its solution. Our paper further contributes to this discussion by introducing an integer programming model that relies on only a polynomial number of variables and constraints that can be solved by CPLEX \cite{CPLEXmanual} for problem sizes that are realistic in applications. 

\section{Problem Formulation}
%=========================================================
For a given Boolean matrix  $X\in\B^{n\times m}$ and integer parameter $k$, we next describe how to construct two Boolean matrices $C\in\B^{n\times k}$ and $R\in\B^{k\times m}$ so as to minimize the approximation error 
$\|X-C\circ R\|_F^2=\sum_{i\in N,j\in M}|x_{i,j}-C_i\circ R_j|$, 
where $C_i$ and $R_j$ denote the $i$-{th} row of $C$ and $j$-th column of $R$, and $N:=\{1,\ldots,n\}$, $M:=\{1,\ldots,m\}$ and $K:=\{1,\ldots,k\}$. In the IP formulation below we denote the McCormick envelope \cite{mccormick:76} of $a,b\in[0,1]$ by $MC(a,b):=\{y\in\R\::\: a\ge y,~b\ge y,~y\ge a+b-1,~y\ge 0\}\subseteq[0,1]$. Note that when $a,b\in\B$, then $MC(a,b)$ only contains the point $ab\in\B$, allowing us to express the nonlinear relationship $y=ab$ in terms of linear constraints only. Optimal factors $C, R$ may now be computed by solving the following binary integer program,
\begin{align}
\text{(BP)}\quad\min_{\xi,z,c,r,y}\,\sum_{i=1}^n \sum_{j=1}^m \xi_{i,j},&\nonumber\\
\text{s.t. }\quad x_{i,j}-z_{i,j}&\leq\xi_{i,j},\qquad z_{i,j}-x_{i,j}\leq\xi_{i,j},&\forall i\in N;\,j\in M,\label{nulla1}\\
z_{i,j}&\leq\sum_{\ell=1}^k y_{i,\ell,j},\qquad y_{i,\ell,j}\leq  z_{i,j},&\forall i\in N;\,j\in M;\,\ell\in K,\label{egy}\\
y_{i,\ell,j}&\in MC(c_{i,\ell},r_{\ell,j}),&\forall i\in N;\,j\in M;\,\ell\in K,\label{form:MC}\\
\xi_{i,j},\;z_{i,j},\;c_{i,\ell},\;r_{\ell,j},\;y_{i,\ell,j}&\in\B,&\forall i\in N;\,j\in M;\,\ell\in K, \label{form:integrality}
\end{align}
where variables $c_{i,\ell}$ and $r_{\ell,j}$ denote the coefficients of $C$ and $R$ respectively, variables $z_{i,j}$ the coefficients of $Z=C\circ R$, $\xi_{i,j}$ the elements of $\Xi=|X-Z|$, and $y_{i,\ell,j}$ the product of the variables $c_{i,\ell}$ and $r_{\ell,j}$. Constraints \eqref{form:integrality} ensure that all variables take $\{0,1\}$ values in any feasible solution. Constraints \eqref{nulla1} imply that $|x_{i,j}-z_{i,j}|\leq\xi_{i,j}$ for all $i\in N,\,j\in M$ and due to the objective function, it is easy to see that $|x_{i,j}-z_{i,j}|=\xi_{i,j}$ in any optimal solution, as desired.
Furthermore, any integral solution satisfies $y_{i,\ell,j}=c_{i,\ell}r_{\ell,j}$ for all $i\in N,\,j\in M,\,\ell\in K$ due to constraints \eqref{form:MC} and therefore constraints \eqref{egy} imply that $z_{i,j}=\min\{1,\sum_{\ell=1}^k y_{i,\ell,j}\}=\min\{1,\sum_{\ell=1}^k c_{i,\ell}r_{\ell,j}\}$, as desired.

\subsection{Improved Formulation}
Note that (BP) uses $O(kmn)$ constraints and variables. To the best of our knowledge, previous IP models for the Boolean rank-$k$ approximation problem from the literature all required an exponential number of constraints 
\cite{Lu:2008}. We remark that the second set of constraints \eqref{egy}, $y_{i,\ell,j}\leq  z_{i,j}$, may be summed and  replaced by constraints $\sum_{\ell} y_{i,\ell,j}\leq  k z_{i,j}$ $(i\in N, j\in M)$ without changing the feasible set. Even though this formulation has fewer constraints, we found it to be less effective in computations because it caused greater branching in the code. However, the formulation (BP) can be significantly improved in other ways to make more efficacious use of the computational power of commercial solvers such as CPLEX \cite{CPLEXmanual}. We next describe these ideas.

\subsubsection{Relaxation of Integrality Constraints}
First note that, due to the nature of McCormick envelopes,  the variables $y_{i,\ell,j}$ are guaranteed to take integer values when  $c_{i,\ell},r_{\ell,j}\in\B$. Consequently, the integrality constraint on $y$ may be relaxed. Further note that if all $y$ variables are integral, then for any fixed $i\in N,\,j\in M$, \eqref{egy} either implies that $z_{i,j}=0$ (if $y_{i,\ell,j}=0$ for all $\ell\in K$) or that $z_{i,j}=1$ (if at least one $y_{i,\ell,j}=1$ for some $\ell\in K$). Therefore, $z$-variables may also be treated as continuous. Finally, since $|x_{i,j}-z_{i,j}|=\xi_{i,j}$ holds at all optimal solutions, the integrality of the $\xi$ variables can also be relaxed. Consequently, only the $r$ and $c$ variables need to be declared integral which leads to a mixed integer programming formulation with $k(n+m)$ binary variables only, which is obtained by replacing constraints \eqref{form:integrality} by 
\begin{equation}\label{form:integrality relaxed}
\xi_{i,j},\;z_{i,j},\;y_{i,\ell,j}\in [0,1],\;
c_{i,\ell},\;r_{\ell,j}\in\B, \quad \forall i\in N;\,j\in M;\,\ell\in K.
\end{equation}

\subsubsection{Deleting Redundant Constraints}
Note that for any $i\in N$ and $j\in M$ the variable $z_{i,j}$ takes a binary value in any feasible solution, and since the input data parameter $x_{i,j}$ is binary as well, one of the two constraints \eqref{nulla1} is redundant. Specifically, we may replace \eqref{nulla1} by 
\begin{equation}\label{nulla1 reduced x=0}
\begin{cases}z_{i,j}=\xi_{i,j},\quad&\text{if } x_{i,j}=0,\\
1-z_{i,j}=\xi_{i,j},\quad&\text{if } x_{i,j}=1.
\end{cases}
\end{equation}
This in turn implies that only one of the constraints \eqref{egy} that involve $z_{i,j}$ is non-redundant, and that \eqref{egy} may be replaced by 
\begin{equation}\label{egy reduced}
\begin{cases}
y_{i,\ell,j}\leq  z_{i,j},\quad&\text{if }x_{i,j}=0,\\
z_{i,j}\leq\sum_{\ell=1}^k y_{i,\ell,j},\quad&\text{if } x_{i,j}=1.
\end{cases}
\end{equation}
Using the same reasoning, one finds that half the constraints that define $MC(c_{i,\ell},r_{\ell,j})$ are redundant, so that 
\eqref{form:MC} may be replaced by 
\begin{equation}\label{form:MC reduced}
\begin{cases}y_{i,\ell,j}\geq r_{\ell,j}+c_{i,\ell}-1,\;y_{i,\ell,j}\geq 0,\quad&\text{if } x_{i,j}=0,\\
r_{\ell,j}\geq y_{i,\ell,j},\;c_{i,\ell}\geq y_{i,\ell,j},\quad&\text{if } x_{i,j}=1.
\end{cases}
\end{equation}
Consequently, approximately half of the constraints in the original formulation can be deleted.

\subsubsection{Preprocessing the Input Data} 
In practice, the matrix $X\in\B^{n\times m}$ of input data may contain rows (or columns) of all zeros. 
Deleting these rows (or columns) leads to an equivalent problem whose solution $C$ and $R$ can easily be 
translated to a solution for the original problem by inserting a row of zeros to $C$ (respectively a column of zeros to $R$) in the corresponding place. In addition, $X$ may contain duplicate rows (or columns). In this case it is sufficient 
to keep only one copy of each, solve the problem on the reduced data, and to reinsert the relevant copies of rows of $C$ (respectively columns of $R$) in the optimal factors $C$ and $R$ found for the reduced data. In order for this process to yield the correct result, the objective function of the reduced problem must correspond to the approximation error of the original problem. Therefore, if row $i$ of $X$ is repeated $\alpha_i$ times, and column $j$ is repeated $\beta_j$ times, then the variable $\xi_{i,j}$ that corresponds to the remaining copy of these rows and columns must be multiplied by $\alpha_i \beta_j$ in the objective function of the reduced problem. 

\section{Computational Experiments}

In this section we report numerical tests of the model (BP) on artificial and real-world datasets. The model was solved by CPLEX \cite{CPLEXmanual} in each instance on a 2015 MacBook Pro with a 3.1 GHz Intel Core i7 processor and 16 GB of memory. The experiments with artificial data reveal that the improved formulation leads to significant speed-up.

\begin{figure}
\begin{center}
\scalebox{0.3}{\includegraphics{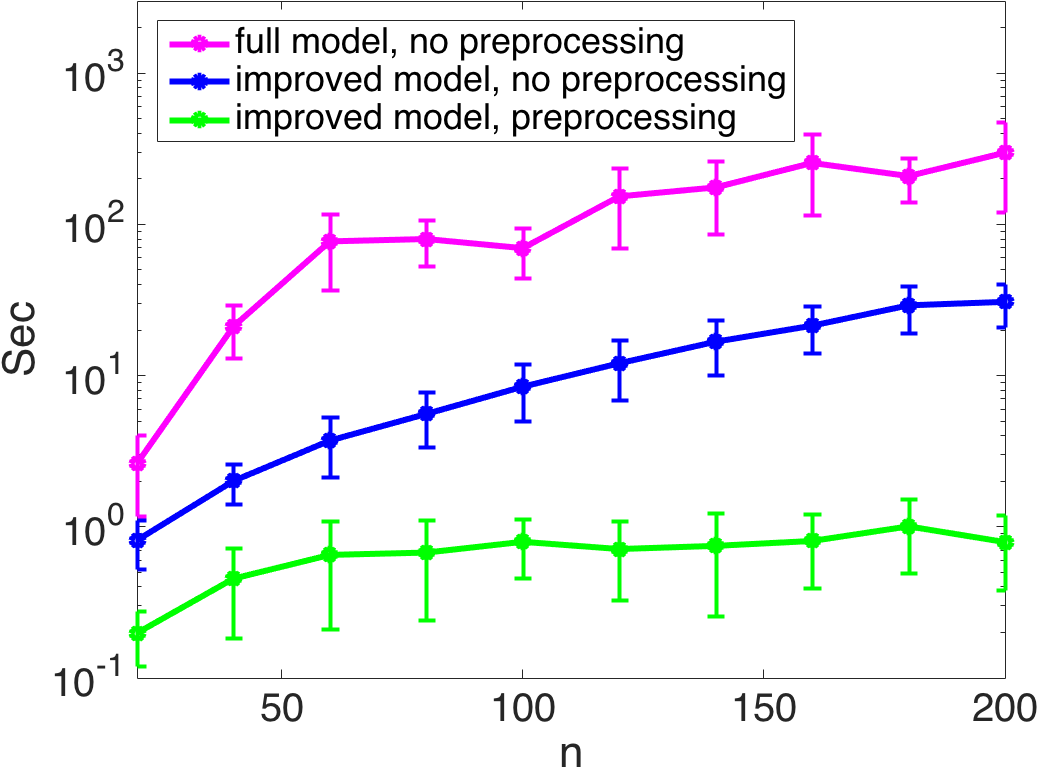}}
\caption{Synthetic exact rank-$5$ data, $m=20$.}\label{figa}
\end{center}
\end{figure}

\subsection{Artificial Datasets} 
\begin{figure}
\begin{center}
\scalebox{0.3}{\includegraphics{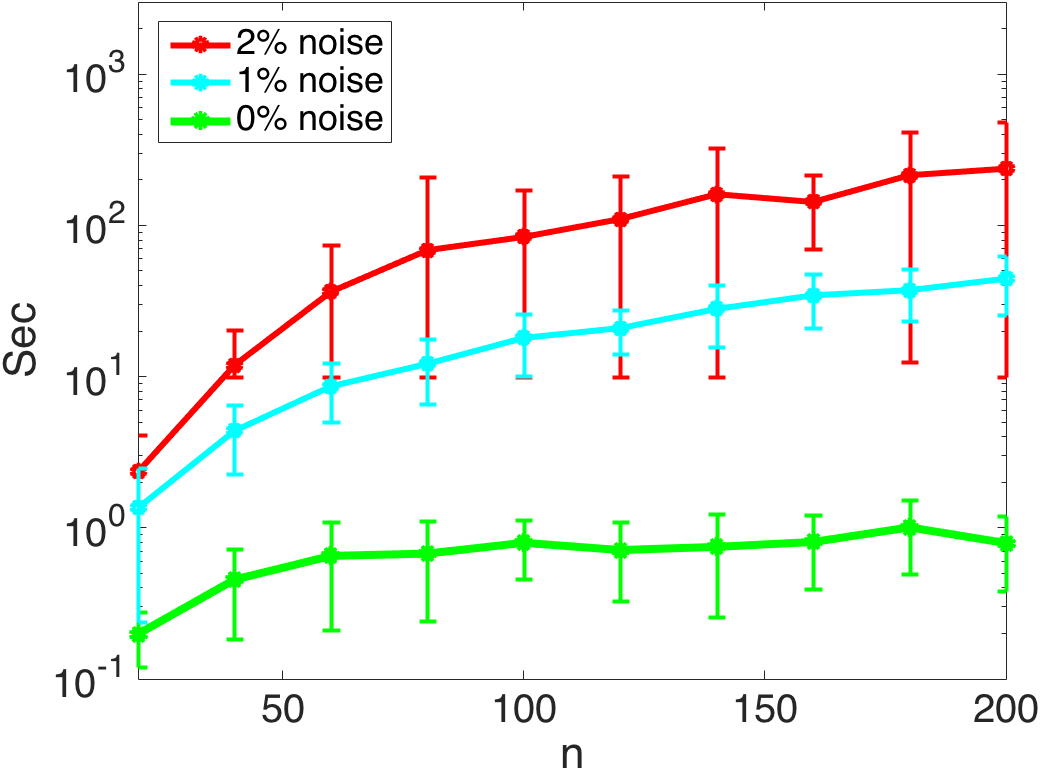}}
\caption{Synthetic noisy rank-$5$ data, $m=20$.}\label{figb}
\end{center}
\end{figure}

Artificial datasets were generated by sampling matrices $C\in\B^{n\times \kappa}$ and $R\in\B^{\kappa\times m}$ with i.i.d.\ random coefficients for fixed parameters $n,m,\kappa$. The data matrix 
$X$ was computed by forming the Boolean product $C\circ R$, which is a matrix with exact Boolean rank $\kappa$, 
and by randomly flipping $\mu\%$ of the coefficients. The sparsity of $X$ was controlled so that 
$x_{i,j}=1$ with probability $1/2$. The results of Figure \ref{figb} show how noise in the data affects the complexity of the problem. Each experiment was repeated $50$ times, and the plot shows averaged running times and error bars. We observe that the complexity of the problem grows rapidly with the level of noise, an effect that occurs partly, but not solely, because the preprocessing step is less powerful at reducing the dimension of noisy matrices. This is confirmed by Figure \ref{figa}, which shows the effect of the improved formulation and the preprocessing steps. Each randomly generated input was used in the full model, the improved model, and the improved model with a preprocessing step, and the running times were averaged over $50$ repeats of the experiment.

%Therefore, the generated matrix $X$ is at most $0.0\mu \cdot nm$ distance away from a Boolean rank-$\kappa$ matrix measured in the squared Frobenius norm.

\subsection{Real-World Datasets}
\begin{figure}
\begin{center}
\scalebox{0.3}{\includegraphics{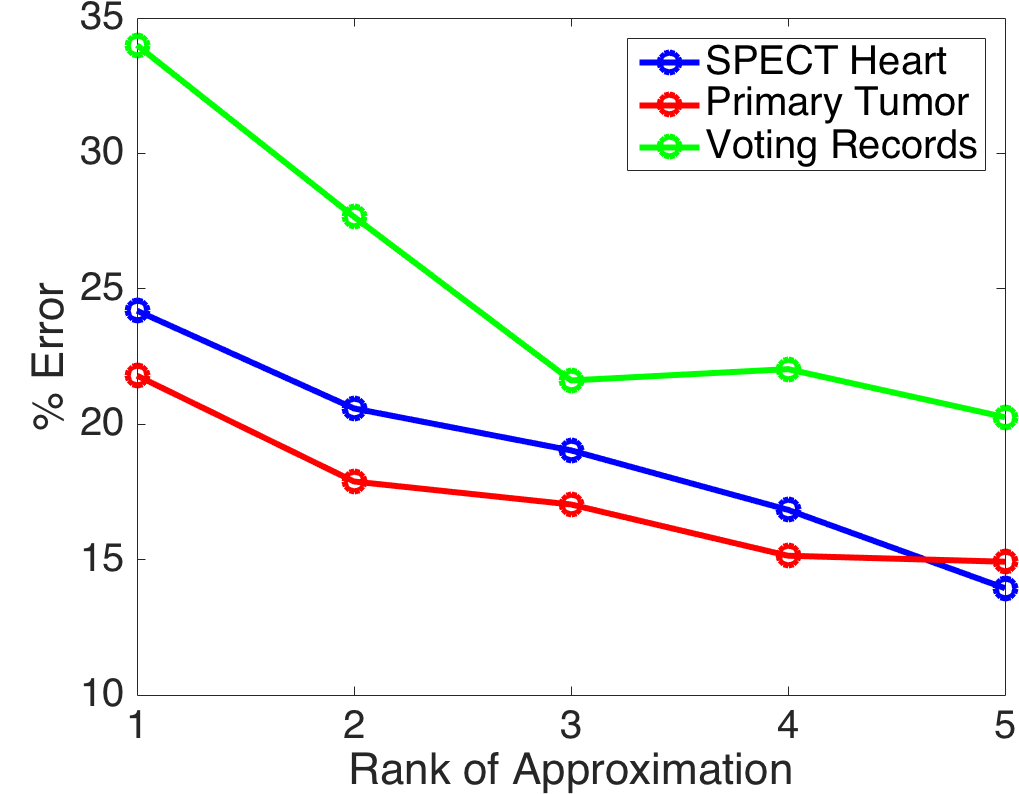}}
\caption{Percentage of approximation error as a function of the approximating rank.}\label{figc}
\end{center}
\end{figure}

%SPECT Heart Data Set
The \textit{SPECT Heart dataset} \cite{SPECT} describes cardiac Single Proton Emission Computed Tomography images of $267$ patients by $22$ binary feature patterns, providing us a Boolean data matrix of dimension $267\times 22$.
%Primary Tumor Data Set
The \textit{Primary Tumor dataset} \cite{TUMOR} contains observations of $17$ categorical variables for $339$ patients. $4$ of the variables were non-Boolean and were converted to $11$ Boolean variables, resulting in an all Boolean representation of the data matrix in dimension $339\times 24$.
%Voting Data Set
The \textit{1984 United States Congressional Voting Records dataset} \cite{VOTING} includes votes for each of the U.S. House of Representatives Congressmen on the $16$ key votes identified by the CQA. The $16$ categorical variables taking values of ``voted for'', ``voted against'' or ``did not vote'', are converted into $32$ Boolean variables. The resulting Boolean data matrix is of dimension $435\times 32$. Rank-$k$ approximations were computed for all three datasets with $k=1,\dots,5$ separately. Each run was limited to a budget of 2h, after which the incumbent solution was used as an approximation and the error reported in Figure \ref{figc}. We observe that even a suboptimal rank-$5$ approximation correctly reconstructs at least $80\%$ of the data in each case. If the problems were solved to optimality, the approximation error would decrease in $k$, but since the figure shows the approximation errors of suboptimal solutions the curves can be non-monotonic. 
% adjust 43 in \begin{wrapfigure}[43]{r}{.4\textwidth} to finish wrapping the space at the correct line

\section{Conclusions}

To the best of our knowledge, the MIP formulation of the optimal low-rank Boolean matrix approximation problem discussed in this paper is the first model that relies on only polynomially many variables and constraints and 
constitutes the first exact method that is viable for realistic problem sizes. Our preliminary computational experiments suggest that our technique is applicable to real world data sets that were hitherto only approachable via heuristics \cite{Miettinen}. 
%
%%\newpage
%\clearpage
%{\small
%  \bibliographystyle{alpha}
%  \bibliography{references}
%}

\newcommand{\etalchar}[1]{$^{#1}$}

\end{document}